\pdfoutput=1

\documentclass[11pt]{article}

\usepackage{naacl2021}

\usepackage{times}
\usepackage{latexsym}
\usepackage{graphicx}
\usepackage{comment}
\usepackage{hyperref}
\usepackage{adjustbox}
\usepackage{mathtools}
\usepackage{array}
\usepackage{bm}
\usepackage{makecell,booktabs}
\usepackage{stackengine}

\usepackage[T1]{fontenc}

\usepackage[utf8]{inputenc}

\usepackage{microtype}

\usepackage{color}

\title{Exploring Low-Cost Transformer Model Compression for Large-Scale Commercial Reply Suggestions}
\author{Vaishnavi Shrivastava\textsuperscript{\rm$\ddagger$}\Thanks{ Equal contribution.}  , Radhika Gaonkar\textsuperscript{\rm$\ddagger$}\footnotemark[1] ,  Shashank Gupta\textsuperscript{\rm$\ddagger$}\footnotemark[1],  Abhishek Jha\textsuperscript{\rm$\spadesuit$}\Thanks{ Work was done while at Microsoft.} \\\\
        \textsuperscript{\rm$\ddagger$}\textbf{Microsoft Search, Assistant and Intelligence} \\
        \texttt{\{vashri, ragaonka, shagup\}@microsoft.com} \\\\
        \textsuperscript{\rm$\spadesuit$}\textbf{Stripe} \\
        \texttt{\{abhijha\}@stripe.com}}

\begin{document}

\maketitle
\begin{abstract}
Fine-tuning pre-trained language models improves the quality of commercial reply suggestion systems, but at the cost of unsustainable training times. Popular training time reduction approaches are resource intensive, thus we explore low-cost model compression techniques like Layer Dropping and Layer Freezing. We demonstrate the efficacy of these techniques in large-data scenarios, enabling the training time reduction for a commercial email reply suggestion system by $42\%$, without affecting the model relevance or user engagement. We further study the robustness of these techniques to pre-trained model and dataset size ablation, and share several insights and recommendations for commercial applications.
\end{abstract}

\section{Introduction}
\label{sec:intro}
Automated reply suggestions have become ubiquitous in many popular email and chat applications, such as Outlook \cite{OutlookSmartReply}, Teams \cite{TeamsSmartReply}, Gmail \cite{GoogleSmartReply}, and LinkedIn \cite{LinkedInSmartReply}. These systems assist the end-users in responding to a message by providing them multiple short contextually-relevant reply suggestions that can be used to reply with a quick click, thus improving their overall productivity. The dominant techniques for these systems model them as a response selection task~\cite{gunasekara2019dstc7} using a bi-encoder matching model that matches the incoming message with a fixed set of pre-selected responses \cite{Henderson2017EfficientNL}. Recently, fine-tuning pre-trained Transformer language model encoders such as BERT \cite{devlin-etal-2019-bert}, T5 \cite{Raffel2020ExploringTL}, or UniLMv2 \cite{Bao2020UniLMv2PL} has shown improvements in model quality \cite{Henderson2020ConveRT}.

However, commercial reply-suggestion systems can muster massive amounts of training data, which leads to large fine-tuning times for these pre-trained models. Furthermore, commercial systems often stack additional rankers on top, wherein the matching model ranker is used earlier in the stack to retrieve relevant responses, and the latter rankers help induce special attributes like diversity \cite{Deb2019DiversifyingRS}, factual correctness, and user  writing style in the responses. For the best model quality, these rankers are often interdependent, and thus a change to a ranker earlier in the stack often requires re-training the entire stack. Additionally, strict compliance with data privacy laws like GDPR \cite{gdpr} requires changing the underlying training dataset every few weeks, requiring all the rankers to be re-trained. These factors make reducing the fine-tuning times of the matching model vital for efficient and continuous development of commercial reply-suggestion systems.

Popular approaches for reducing the training times involve compressing the pre-trained model size using techniques like Distillation \cite{Hinton2015DistillingTK}. Such techniques, however, are resource intensive, and require additional training. In this work, we explore low-cost model compression solutions like Layer Dropping and Layer Freezing. We demonstrate that these low-cost techniques are highly effective in large-data scenarios, helping reduce the training times of a commercial email reply suggestion system by over $42\%$. We also show that these techniques can be combined together into a hybrid model for additional gains. We further study the robustness and efficacy of these compression techniques with ablations in the pre-trained model and the training dataset size, and share our findings and recommendations. We hope that these findings will help inform the experimentation and design of commercial applications even beyond reply suggestion scenarios. \\

Specifically, our findings for \textbf{large-data scenarios} can be summarized as:

\begin{enumerate}
\item For the same desired amount of training time reduction, compressing the model and training it on a large dataset preserves the relevance better than training a large model on a downsampled dataset (Section~\ref{ssec:dataset_res}).
\item Layer Dropping and Layer Freezing are competitive low-cost model compression techniques, and thus should be prioritized over computationally expensive techniques like Distillation (Sections~\ref{ssec:reduction_res} and~\ref{ssec:freezing_res}).
\item A randomly initialized model with sufficient capacity might be able to match the relevance of a pre-trained initialized model at large dataset scale, however, the usage of a pre-trained model is still recommended since it compresses better and can lead to reduced training times (Section~\ref{sec:elr}).
\item However, further improving the pre-trained model through its domain-adaptation doesn't offer any additional relevance or compression advantages over the vanilla pre-trained model at large dataset scales (Section~\ref{sec:elr}). We thus advise caution before prioritizing improvements to the underlying pre-trained model for task-specific relevance and training time gains.
\end{enumerate}

Similarly, our findings for \textbf{low-data scenarios} can be summarized as:
\begin{enumerate}
\item In keeping with the literature, we found that initialization with a pre-trained model  is vastly superior to a randomly initialized model of the same capacity. Furthermore, we found that domain-adaptation of the pre-trained model provides additional relevance gains (Section~\ref{sec:dataset_ablation}). We thus recommend leveraging a pre-trained model and trying out its improvements (for instance its domain-adaptation) in low-data settings.
\item We discovered that the efficacy of model compression techniques can be a function of the size of the fine-tuning dataset. Specifically, we found that Layer Dropping is only competitive on large fine-tuning datasets, whereas the performance of Layer Freezing is agnostic to the dataset size. We thus recommend Layer Freezing over more complex model compression techniques across dataset sizes (Section~\ref{sec:dataset_ablation}).
\end{enumerate}

We start in section~\ref{sec:model} by providing a brief background of our reply suggestion system and the compression techniques being pursued, followed by a brief overview of the related work in section~\ref{sec:rel}. We discuss our experimental setup in section~\ref{sec:exp_setup}, and present the training time reduction results on a large dataset in section~\ref{sec:results}. We then study the impact of the pre-trained model, and the dataset size in sections~\ref{sec:elr} and~\ref{sec:dataset_ablation} respectively. We discuss future work in section~\ref{sec:future}, and finally conclude in section~\ref{sec:conclusion}.

\section{Background}
\label{sec:model}
\subsection{Reply Suggestion System}
\label{ssec:matching_desc}
\begin{figure}
\centering
\includegraphics{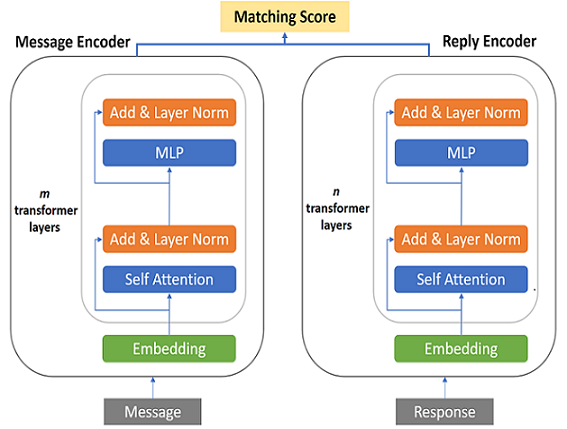}
\caption{Matching model architecture consisting of a bi-encoder over the message-reply pairs}
\label{fig:matching_l1}
\end{figure}

We model suggesting replies to emails as a retrieval problem from a fixed set of pre-selected candidate responses (a Response Set). Given an email dataset consisting of message-response (MR) pairs, we use a bi-encoder matching architecture to encode the message and response using distinct transformer encoders ~\cite{Vaswani2017AttentionIA} (as in Figure \ref{fig:matching_l1}), and then train the model through a symmetric matching loss \cite{Deb2019DiversifyingRS} minimization objective between the message and response encodings.

For inference, we rank the responses from the response set using a combination of their matching score with the message (as provided by the bi-encoder), and an additional bias term to promote responses with high pre-computed message-independent LM scores (language model scores). The bias term balances the matching model's tendency to surface long and highly specific responses by promoting generic responses, and is a hyper-parameter that we tune. We then use lexical clustering to select $3$ diverse responses from the top ranked responses, such that each of the selected responses comes from a different cluster. In our system, we pre-compute and cache the response set encodings, which makes computing the encoding of the incoming message our online serving bottleneck, and incentivizes keeping the message encoder compact. For instance, our baseline system uses $12$ transformer layers to encode the response, and just $6$ to encode the message. 
This formulation is similar to the work of Henderson et al \cite{Henderson2017EfficientNL}.

Our response set consists of a set of responses from our email training dataset that meet a certain frequency threshold are filtered through a set of carefully curated n-grams to prevent them from containing any user information or offensive content. These responses are further reviewed for their quality by a dedicated team of human curators. We use the normalized log frequency of these responses as their message-independent LM score.

\subsection{Compression Techniques}
\label{ssec:compression_tech}
Literature has shown that pre-trained models are heavily overparameterized, with possible redundancies between layers ~\cite{2020PoorMansBERT}, and only some subset of the layers needing to be fine-tuned \cite{Lee2019ElsaFreezing}. Thus, several model compression techniques like Distillation, Pruning, and Quantization have been proposed to curb the training times. In this work we investigate the effectiveness of low-cost techniques like Layer Dropping and Freezing.

\textbf{Layer Dropping:} 
Layer Dropping involves heuristically pruning entire layers from the message and response encoders, and fine-tuning the resulting model. Fewer layers lead to reduced training times and online latencies. Dropping layers also gives the flexibility of selecting which layers to retain from the pre-trained model, and since different layers can capture different linguistic information \cite{tenney-etal-2019-bert, Clark2019BertAttention}, end-performance can vary based on the layers selected \cite{2020PoorMansBERT}.

\textbf{Layer Freezing:}
Layer Freezing involves not updating the weights of a bottom few encoder layers during fine-tuning, since the bottom layers usually capture task-agnostic information \cite{Lee2019ElsaFreezing}. This helps reduce the training time by optimizing the backward-pass, however unlike Layer Dropping, it doesn't provide any inference time reduction.

Note that Layer Dropping and Freezing can additionally be combined together into a \textbf{Hybrid Model} to reap the benefits of both approaches.
\subsection{Pre-trained Model}
\label{ssec:elr_background}
Pre-trained models like BERT \cite{devlin-etal-2019-bert} are trained on large amounts of text from Wikipedia and BookCorpus ~\cite{Zhu2015AligningBA}, and thus provide useful representations for many domains. Recently, adaptation of these models to the target domain through continued pre-training ~\cite{Gururangan2020DontSP} or training from scratch ~\cite{gu2020domainspecific} has shown to further improve their representations. In this work we study if the domain-adaptation of the pre-trained model leads to better model compression as well.

\section{Related Work}
\label{sec:rel}
Many model compression techniques have been proposed to improve the fine-tuning times of Transformer-based pre-trained models. Distillation \cite{Sanh2019DistilBERTAD} is a widely used technique where a large pre-trained teacher model is distilled into a smaller student model using a distillation loss. However, pre-trained model distillation can be computationally expensive, thus making it unsuitable for catering to the ad-hoc model size reduction needs of various scenarios. Our work instead explores simple model compression techniques like Layer Dropping and Layer Freezing \cite{2020PoorMansBERT, Lee2019ElsaFreezing} that incur no additional computation overhead, and (unlike Distillation) seamlessly support ad-hoc model size reductions by not requiring the pre-trained model to be first distilled down to the desired model size.

Our Layer Dropping technique can also be considered to be a special case of existing techniques like weight or attention head pruning \cite{raganato-etal-2020-fixed, gordon-etal-2020-compressing}, wherein we heuristically prune entire layers instead of pruning specific weights or attention heads.

Most model compression techniques in literature have been evaluated on public datasets with only modest amounts of fine-tuning data, which is not representative of the scale of fine-tuning datasets available for commercial applications. In this work, we take the first step by evaluating simple model compression techniques on large amounts of fine-tuning data, and study their efficacy with variations in the size of the fine-tuning dataset.

Furthermore, our experiments with the domain-adapted pre-trained model are motivated by their recent success in the literature \cite{Gururangan2020DontSP}, wherein they have been found to perform better than their vanilla counterparts. However, similar to model compression techniques, the domain-adapted pre-trained models have only been evaluated on tasks with small amounts of fine-tuning data. Our work complements the existing work by experimenting with domain-adapted pre-trained models with varying amount of fine-tuning dataset. We additionally study if domain-adapted pre-trained models offer any model compression advantages.

\section{Experimental Setup}
\label{sec:exp_setup}
\textbf{Notation:} We use $M_xR_y$ to denote a base matching model with $x$ and $y$ transformer blocks in the message and response encoders respectively. Furthermore, whenever we freeze some layers of the $M_xR_y$ base model, we will embellish the notation as $(M_xR_y, fm_ir_j)$ to denote that the subword embeddings and the bottom-most $i$ message and $j$ response encoder layers have been frozen during training. We will represent freezing just the subword embeddings on both encoders as $(M_xR_y, f_{emb})$.
For Layer Freezing configurations, whenever the base model is clear from the context, we will drop the base model notation and will refer to the model with just the freezing notation $fm_ir_j$.

\textbf{Baseline Model:} We use a BERT-base bi-encoder architecture \cite{devlin-etal-2019-bert}, with $6$ Transformer layers in the message encoder\footnote{using 12 layers didn't result in any relevance gains} and $12$ in the response (\bm{$M_6R_{12}$}) encoder to keep our serving latencies small. We don't share weights between the encoders during training, use the CLS token's output for scoring responses, use the same segment id (of $0$) in both the encoders\footnote{we noticed no gains with using different segment ids}, a wordpiece uncased vocab of size $30k$, and initialize the weights of encoders with a pre-trained BERT base-like model, called Turing-NLR (\textbf{TNLR}), from Microsoft Turing\footnote{https://turing.microsoft.com/}.

\textbf{Training:} Unless otherwise stated, we train on roughly $60$ million message-reply (MR) pairs extracted from the mailboxes of users of a popular email client. We adhere to strict user privacy policies, and use state-of-the-art privacy preserving technologies. The user data is always kept encrypted, and the model training is carried in a secure environment that doesn't allow any viewing of the contents of the mailboxes. We use $16$ Nvidia V100 GPUs with InfiniBand connectivity for distributed training. We use gradient accumulation and mixed precision training\footnote{https://nvidia.github.io/apex/amp.html} to speed up the training, and use the Adam optimizer with a scheduled learning rate, involving a linear warmup followed by an exponential decay. We train for a max of $20$ epochs\footnote{$20$ epochs were sufficient for our models to converge} and use validation on $20k$ instances to select the best checkpoint.

\textbf{Evaluation:} The objective of this work is to reduce the training times for our baseline \bm{$M_6R_{12}$} while maintaining the relevance of the model. 

For measuring training time improvements, we use Wall-Clock (WC) time, which we define as the time taken to obtain the best model checkpoint. This depends on the epoch time and the rate of convergence of the model.

For measuring the model relevance offline, we compute a w-Rouge measure on a set of $500k$ email-reply pairs. We define  w-Rouge as a weighted version of the Rouge F-measure \cite{lin-2004-rouge} that can measure similarity between a user's actual (golden) response and multiple predicted responses (a response block). For each MR pair instance, we first compute the maximum Rouge-n score between the golden response ($r^*$) and each of the $3$ predicted responses ($r_i$). We define it as Rouge-n\textsubscript{max} (equation~\ref{eq:max_rouge}), where $n$ is the length of the ngram. We then take a weighted average ($w_n$ being the weight) of these Rouge-n\textsubscript{max} scores across n-grams of length 1, 2, and 3 as in equation \ref{eq:weighted_rouge}. We ultimately macro-average these w-Rouge scores across all test instances.
    \begin{equation}
    \label{eq:max_rouge}
    \begin{aligned}
        \operatorname{{Rouge-n_{max}}} = \max_{i\in\{1, 3\}} \operatorname{Rouge-n_{(r^*, r_{i})}}
    \end{aligned}
    \end{equation}

    \begin{equation}
    \label{eq:weighted_rouge}
    \begin{aligned}
        \operatorname{w-Rouge} = {\sum_{n=1}^{3}} & w_{n} \times \operatorname{Rouge-n_{max}}
    \end{aligned}
    \end{equation}
We use w-Rouge over other IR metrics like Mean Reciprocal Rank, since it allows partial overlap, and has demonstrated a higher correlation with our user engagement metrics. We run two-sided T-tests, and only report statistically significant changes where \textit{p-value $\leq 0.05$}.

The ultimate evaluation of the impact on model relevance is through end-user engagement, for which we divide users into treatment and control groups and conduct experiments with a small percentage of online user traffic. We use user-averaged click-through-rate (CTR) as our engagement metric, and only report statistically significant changes to the CTR \textit{(p-value $\leq 0.05$)}. We thus seek a model with significantly smaller WC time and no drop in online CTR compared to the baseline $M_6R_{12}$.

However, since running online experiments is costly and affects the user experience, we run them only for the most promising models with no w-Rouge drop, and use w-Rouge to make inferences for the rest.

\textbf{Pre-trained Models:} We experiment with $3$ pre-trained models -- Random (i.e. no pre-training), TNLR, and D-TNLR. 

TNLR is Project Turing's BERT-base equivalent which is trained from scratch similar to BERT, but with the addition of phrase-level masking and replacement of NSP loss with a `Sentence Order Prediction (SOP)' loss. 
We use TNLR since it has demonstrated improvements over BERT-base on the GLUE dev set \cite{wang-etal-2018-glue}.

D-TNLR is the domain-adapted version of TNLR. It is trained from scratch like TNLR on a combination of public datasets and domain-specific commercial email datasets. All training was done in a fully-compliant eyes-off manner, with data given the proper privacy permissions to be used in training and validation. D-TNLR upsamples the email corpus during training, interleaves batches between datasets, and has a performance competitive to TNLR on the GLUE dev set.

\section{Results \& Discussion}
\label{sec:results}
\textbf{BERT vs TNLR:} We first fine-tune and compare the public BERT-base\footnote{https://github.com/huggingface/transformers} and TNLR baseline models for our task. We see in Table~\ref{table:tnlr_bert} that in addition to similar GLUE scores, the pre-trained models are identical for our task as well (we use $\dagger$ to represent stat-sig. changes throughout). We can thus expect our findings from TNLR to generalize to BERT as well.

\begin{table}
\centering
\begin{tabular}{ll}
\hline
\textbf{Initialization} & \textbf{w-Rouge}\\
\hline
BERT & $0.07232$\\
TNLR & $0.07261$\\
\hline
\end{tabular}
\caption{Comparison of TNLR \& BERT on $M_6R_{12}$}
\label{table:tnlr_bert}
\end{table}
\subsection{Dataset Downsampling}
\label{ssec:dataset_res}
We evaluate downsampling the training dataset as a viable training time reduction technique. Table~\ref{table:baseline_dataset} shows that a $90\%$ sample ($54M$) shows no drop in w-Rouge, but yields only a modest time reduction ($16\%$). However, when we sample more aggressively ($\leq40\%$ or $\leq25M$), we see a statistically significant $\geq1.9\%$ drop in w-Rouge. This shows that we can't rely on downsampling for our desired amounts of training time reduction, while maintaining relevance.
\begin{table}
\centering
\begin{tabular}{lll}
\hline
\textbf{Dataset size} & \textbf{Time} & \textbf{w-Rouge}\\
\hline
$60M$ & $55.8$hrs & $0.07261$ \\
\hline
\bm{$54M$} & \bm{$46.8$}\textbf{hrs} & \bm{$0.07227$} \\
$25M$ & $22.8$hrs & $0.07123^\dagger$ \\
$5M$ & $25.7$hrs & $0.06919^\dagger$ \\
$100k$ & $9.6$hrs & $0.05933^\dagger$ \\
\hline
\end{tabular}
\caption{Impact of Dataset Downsampling (\#MR pairs)}
\label{table:baseline_dataset}
\end{table}
\subsection{Layer Dropping}
\label{ssec:reduction_res}
We first quantify the impact of the position of the layers being retained from the pre-trained model. Towards this goal, we benchmark retaining different layers ($0$ indexed) in the message encoder of a $M_3R_{12}$ model compressed with Layer Dropping -- \textit{Bottom-$3$} (Layers $0, 1, 2$), \textit{Top-$3$} (Layers $9, 10, 11$), \textit{Even-$3$} (Layers $0, 6, 10$), and \textit{Odd-$3$} (Layers $1, 7, 11$). Table~\ref{table:layer_selection} shows that different layer selections are comparable, and thus we consistently use the \textit{Odd} layer selection in the rest of the Layer Dropping experiments, since it yields slightly better performance.

\begin{table}
\centering
\begin{tabular}{ll}
\hline
\textbf{Layers} & \textbf{w-Rouge}\\
\hline
Bottom-3 & $0.07221$ \\
Top-3 & $0.07232$ \\
Even-3 & $0.07216$ \\
Odd-3 & $0.07252$ \\
\hline
\end{tabular}
\caption{Impact of Layer Selection on $M_3R_{12}$}
\label{table:layer_selection}
\end{table}
\textbf{Offline Results:} We experimented with various degrees of Layer Dropping in the message and response encoders, and found the WC times to go down as we removed more layers (Table~\ref{table:reduction_tnlr_full}). We found several configurations with no drop in w-Rouge, with \bm{$M_3R_3$} standing out with \textbf{48\% fewer parameters}, and a \textbf{42.3\% training time reduction}. Further removal of layers, as in with $M_2R_2$, led to a substantial $1.5\%$ drop in w-Rouge.

\begin{table}
\centering
\begin{tabular}{llll}
\hline
\textbf{Model} & \textbf{\#Params} & \textbf{Time} & \textbf{w-Rouge}\\
\hline
$M_6R_{12}$ & $176M$ & $55.8$hrs & $0.07261$ \\
\hline
$M_3R_{12}$ & $155M$ & $46.2$hrs & $0.07252$ \\
$M_3R_6$ & $113M$ & $36.3$hrs  & $0.07238$ \\
\bm{$M_3R_3$} & \bm{$91M$} & \bm{$32.2$}\textbf{hrs} & \bm{$0.07225$} \\
$M_2R_2$ & $77M$ & $27.3$hrs & $0.07148^\dagger$ \\
\hline
\end{tabular}
\caption{Performance of Layer Dropping}
\label{table:reduction_tnlr_full}
\end{table}

\textbf{Online Results:} Seeing favorable offline results for $M_3R_3$, we ran an online experiment comparing this model against the baseline for several weeks using real user traffic, and noticed \textbf{no impact on the CTR} -- indicating an equal preference for the compressed model\footnote{crowd-sourced human judgments showed a similar trend}. This reduction in model size also helped us \textbf{reduce our online serving latencies by over 35\%}, thus saving compute for other parts of the model stack.

\subsection{Layer Freezing}
\label{ssec:freezing_res}
\textbf{Offline Results:} We experimented with freezing message and response encoders of $M_6R_{12}$ to various degrees (Table~\ref{Table:freezing_tnlr_full}), and noticed WC times to drop as we froze more layers. \#Params in (Table~\ref{Table:freezing_tnlr_full}) denotes the number of trainable parameters after Layer Freezing. We noticed a significant drop of $75\%$ in w-Rouge when we froze the entire model, thus showing the importance of task-specific fine-tuning. We found several configurations with no drop on w-Rouge, with \bm{$fm_3r_6$} standing out with \textbf{63\% fewer parameters}, and \textbf{42.7\% reduction in the training time}.

\begin{table}
\centering
\begin{tabular}{llll}
\hline
\textbf{Config} & \textbf{\#Params} & \textbf{Time} & \textbf{w-Rouge}\\
\hline
Unfrozen & 176M & 55.8hrs & $0.07261$ \\
\hline
All frozen & 0 & 0m & $0.01786^\dagger$ \\
$fm_3r_0$ & 131M & 48.8hrs  & $0.07291$ \\
$f_{emb}$ & 129M & 46.3hrs  & $0.07345$ \\
$fm_0r_6$ & 110M & 40.8hrs  & $0.07270$ \\
$fm_3r_3$ & 86M & 36.5hrs  & $0.07246$ \\
\bm{$fm_3r_6$} & \bm{$65M$} & \bm{$32$}\textbf{hrs}  & \bm{$0.07243$} \\
$fm_6r_6$ & 43M & 27.3hrs & $0.05702^\dagger$ \\
$fm_3r_{12}$ & 22M & 25.3hrs & $0.06192^\dagger$ \\
\hline
\end{tabular}
\caption{Performance of Layer Freezing on $M_6R_{12}$}
\label{Table:freezing_tnlr_full}
\end{table}

\textbf{Online Results:} Similar to Layer Dropping, we ran an online experiment comparing $fm_3r_6$ against the baseline for several weeks, and noticed \textbf{no impact on the CTR}. However, since Layer Freezing has no inference time advantages, we noticed no reduction in our online serving latencies.

Overall, these results demonstrate that Layer Dropping and Layer Freezing are effective and comparable training time reduction techniques on large datasets, with Layer Dropping holding an edge due to its inference time advantages as well. These techniques are complementary to each other and thus can be combined into a hybrid model for additional gains as well. Our initial experiments with hybrid models show that it is possible to further freeze the subword embeddings of the $M_3R_3$ model without affecting w-Rouge to obtain a cumulative training time reduction of \bm{$59\%$} (Table~\ref{table:hybrid_tnlr_full}). 
\begin{table}
\centering
\begin{tabular}{lll}
\hline
\textbf{Model} & \textbf{Time} & \textbf{w-Rouge}\\
\hline
$M_6R_{12}$ & $55.8$hrs & $0.07261$ \\
\hline
\bm{$(M_3R_3, f_{emb})$} & \bm{$22.8$}\textbf{hrs} & \bm{$0.07200$} \\
$(M_3R_3, fm_1r_1)$ & $21.5$hrs & $0.07129^\dagger$ \\
\hline
\end{tabular}
\caption{Performance of Hybrid $M_3R_3$ models}
\label{table:hybrid_tnlr_full}
\end{table}

Lastly, we note that our best compressed model $(M_3R_3, f_{emb})$ has a better w-Rouge ($1.1\%$ better) and comparable WC time to training the baseline model $M_6R_{12}$ on the downsampled $25M$ dataset (section~\ref{ssec:dataset_res}). This result suggests that when faced with the need to bring down training times of large models on large datasets, compressing the model and training it on a large dataset preserves the relevance better than training the large model on a downsampled dataset.

\section{Impact of Pre-trained Models}
\label{sec:elr}
\textbf{Random Initialization:} We first benchmark random initialization against TNLR to study the benefits of using a pre-trained model. We see in Table~\ref{table:pretrain_full_compress} that the Random $M_6R_{12}$ baseline performs as well as its TNLR counterpart (p-value: $0.23$). This shows that the relevance benefits of a high-capacity pre-trained model diminish on large datasets\footnote{We also noticed no impact on the convergence.}. However, the pre-trained model is still desirable because of its higher compressibility, since we found Layer Dropping to not work as well for the randomly initialized model (Table~\ref{table:pretrain_full_compress}). Specifically, the Random $M_3R_3$ configuration showed a significant drop in w-Rouge of $1.3\%$ compared to TNLR $M_3R_3$.
\begin{table}
\centering
\begin{tabular}{llll}
\hline
\textbf{Init.} & \bm{$M_6R_{12}$} & \bm{$M_3R_3$} & \bm{$fm_3r_6$}\\
\hline
TNLR & $0.07261$ & $0.07225$ & $0.07243$\\
\hline
Random & $0.07217$ & $0.07129^\dagger$ & --\\
D-TNLR & $0.07261$ & $0.07223$ & $0.07245$\\
\hline
\end{tabular}
\caption{Impact of pre-trained model on various model configurations.}
\label{table:pretrain_full_compress}
\end{table}

\textbf{Domain-Adapted Pre-trained Model:} We then benchmark D-TNLR to check if a better, domain-adapted pre-trained model can help us compress even more. We report in Table~\ref{table:pretrain_full_compress} that the D-TNLR $M_6R_{12}$ baseline performs only as well as its TNLR counterpart. We also see that D-TNLR doesn't provide any compression benefits over TNLR in its Layer Dropping and Layer Freezing configurations. We additionally found D-TNLR to not show any noticeable convergence time benefits as well.

Overall, these results show that even with large fine-tuning datasets, pre-trained models play a role in making compression techniques like Layer Dropping and Freezing competitive. However, improvement to the pre-trained model by adapting it to the target domain does not yield further compression or convergence benefits with such large datasets. Next, we further investigate the generalization of these observations with dataset size variations.

\section{Impact of Dataset Size}
\label{sec:dataset_ablation}
We create $2$ sampled datasets of sizes $5M$ and $100k$\footnote{typical dataset sizes used in response selection literature}, and fine-tune on them to study the impact of dataset size on the effectiveness of Layer Dropping, Freezing, and pre-trained models.

\textbf{Layer Dropping:} We repeat the TNLR Layer Dropping experiments on these smaller datasets (Section~\ref{ssec:reduction_res}), and report in Table~\ref{table:TNLR_reduction_ablation} that Layer Dropping from $M_6R_{12}$ to $M_3R_3$ doesn't work as well on $5M$ and $100k$ datasets, as it leads to $1.80\%$ and $2\%$ drops in w-Rouge respectively. This shows that in addition to the pre-trained model (Section~\ref{sec:elr}), the effectiveness of Layer Dropping also depends on the availability of a large fine-tuning dataset.

\textbf{Layer Freezing:} The corresponding TNLR freezing experiments (Table~\ref{table:TNLR_reduction_ablation}) show however, that Layer Freezing's effectiveness is agnostic to the dataset size. Specifically, $fm_3r_6$ performs as well as the $M_6R_{12}$ baseline on the $5M$ set, and even shows a gain in w-Rouge of $2.5\%$ on the $100k$ set, with the likely explanation here being that freezing potentially helps avoid overfitting on smaller datasets.

These results also show that the preference between Layer Dropping and Freezing changes with the dataset size, with freezing becoming the preferred technique on smaller datasets due to its relevance advantages over dropping.
\begin{table}
\centering
\begin{tabular}{llll}
\hline
\textbf{Config} & \bm{$60M$} & \bm{$5M$} & \bm{$100k$}\\
\hline
$M_6R_{12}$ & $0.07261$ & $0.06919$ & $0.05933$ \\
\hline
$M_3R_3$ & $0.07225$ & $0.06795^\dagger$ & $0.05809^\dagger$ \\
$fm_3r_6$ & $0.07243$ & $0.06926$ & $0.06079^\dagger$ \\
\hline
\end{tabular}
\caption{Comparison of w-Rouge across datasets for various TNLR configs}
\label{table:TNLR_reduction_ablation}
\end{table}

\textbf{Pre-trained Model:} We repeat the experiments of Section~\ref{sec:elr} on the $5M$ set, and show (Table~\ref{table:elr_5M_results}) that compared to TNLR, the Random $M_6R_{12}$ baseline has a drop in w-Rouge of $6\%$. This shows that a pre-trained model is necessary for relevance (and not just for compression) on smaller datasets.

\begin{table}
\centering
\begin{tabular}{llll}
\hline
\textbf{Model} & \textbf{$M_6R_{12}$} & \textbf{$M_3R_3$} & \textbf{$fm_3r_6$}\\
\hline
TNLR & $0.06919$ & $0.06795$ & $0.06926$ \\
\hline
Random & $0.06499^\dagger$ & $0.06447^\dagger$ & N/A \\
D-TNLR & $0.06956$ & $0.06823$ & $0.06978$ \\
\hline
\end{tabular}
\caption{Comparison of pre-trained models on 5M set}
\label{table:elr_5M_results}
\end{table}
However, we also see in Table~\ref{table:elr_5M_results} that even with $5M$ instances, D-TNLR performs only as well as TNLR. We thus repeat the experiments on the $100k$ set, and report in Table~\ref{table:elr_100k_results} that at this scale, the D-TNLR baseline not only gives a $6.4\%$ w-Rouge improvement over the TNLR baseline, but also compresses better than TNLR as its dropping and freezing configurations provide $3.75\%$ and $2.46\%$ gains over TNLR respectively. These results show that adapting the pre-trained model to the target domain is a viable training time reduction technique on small datasets.
\begin{table}
\centering
\begin{tabular}{llll}
\hline
\textbf{Model} & \textbf{$M_6R_{12}$} & \textbf{$M_3R_3$} & \textbf{$fm_3r_6$}\\
\hline
TNLR & $0.05933$ & $0.05809$ & $0.06079$ \\
D-TNLR & $0.06312^\dagger$ & $0.06027^\dagger$ & $0.06348^\dagger$ \\
\hline
\end{tabular}
\caption{Comparison of TNLR and D-TNLR on 100k}
\label{table:elr_100k_results}
\end{table}

\section{Future Work}
\label{sec:future}
Since our recommendations and insights are drawn from our bi-encoder based suggested replies system for a large-scale commercial email service, in the future we will evaluate their generalization to other response selection datasets and architectures. Some early experiments on a large-scale commercial chat client are already showing a similar trend. We will also evaluate the generalization of our findings to other tasks with large amounts of fine-tuning data, primarily the scenarios where implicit labels are readily available such as text prediction. We will also experiment with combining other compression methods such as quantization \cite{Zafrir2019Q8BERTQ8}, weight pruning \cite{gordon-etal-2020-compressing}, and attention head pruning \cite{raganato-etal-2020-fixed, NEURIPS2019_2c601ad9} with our layer dropping and freezing configurations to further reduce our training times.

\section{Conclusion}
\label{sec:conclusion}
We showed that Layer Dropping and Layer Freezing are competitive techniques to compress transformer bi-encoders in commercial, large-data scenarios, and that the presence of a pre-trained model is essential for these compression techniques to work well. In particular, we were able to reduce the training times of a commercial email service's reply-suggestion system by over $42\%$, without affecting the user engagement. The optimal Layer Dropping configuration also led to a $35\%$ improvement in online latency.

Furthermore, we showed that although these strategies perform similarly at large dataset scales, their comparative efficacy changes when dataset size is varied. Specifically, as the size of the fine-tuning data is reduced, while performance can no longer be matched by dropping layers to the same extent, Layer Freezing continues to be competitive, even improving upon the baseline's performance.

We also showed that the advantages of domain-adapted pre-trained models \cite{Gururangan2020DontSP, gu2020domainspecific} diminish in the presence of large fine-tuning datasets. However, as the dataset size is reduced, domain adaptation becomes important again, as it leads to significant relevance and model compression gains. We believe these insights can benefit many such Transformer bi-encoder based models in commercial settings, by helping achieve significant reductions in model training and allowing faster iteration times in deploying model improvements.

\section*{Acknowledgements}
\label{sec:ack}
We gratefully acknowledge the contributions of the entire Suggested Replies team and various partner teams in helping build and maintain different parts of our system upon which this work builds. We thank the Turing and SURGE teams for supporting us with our pre-trained model needs. We also sincerely acknowledge Budhaditya Deb, Ahmed Awadallah, and Milad Shokouhi for their
support and insightful inputs throughout this work.

\bibliography{anthology,custom}
\bibliographystyle{acl_natbib}

\end{document}